\documentclass[10pt,twocolumn,letterpaper]{article}

\usepackage{cvpr}              
\definecolor{cvprblue}{rgb}{0.21,0.49,0.74}
\usepackage[pagebackref,breaklinks,colorlinks,allcolors=cvprblue]{hyperref}
\usepackage{algorithm}
\usepackage{algpseudocode}
\usepackage{xcolor,soul}

\def\paperID{} 
\def\confName{CVPR}
\def\confYear{2026}

\title{\textbf{Anatomy-Aware Unsupervised Detection and Localization of Retinal Abnormalities in Optical Coherence Tomography}}

\author{
Tania Haghighi \quad Sina Gholami \quad Hamed Tabkhi \quad Minhaj Nur Alam\\
University of North Carolina at Charlotte\\
Charlotte, NC, USA\\
}

\begin{document}
\maketitle
\begin{abstract}
Reliable automated analysis of Optical Coherence Tomography (OCT) imaging is crucial for diagnosing retinal disorders but faces a critical barrier: the need for expensive, labor-intensive expert annotations. Supervised deep learning models struggle to generalize across diverse pathologies, imaging devices, and patient populations due to their restricted vocabulary of annotated abnormalities. 

We propose an unsupervised anomaly detection framework that learns the normative distribution of healthy retinal anatomy without lesion annotations, directly addressing annotation efficiency challenges in clinical deployment. Our approach leverages a discrete latent model trained on normal B-scans to capture OCT-specific structural patterns. To enhance clinical robustness, we incorporate retinal layer-aware supervision and structured triplet learning to separate healthy from pathological representations, improving model reliability across varied imaging conditions. During inference, anomalies are detected and localized via reconstruction discrepancies, enabling both image and pixel-level identification without requiring disease-specific labels. On the Kermany dataset (AUROC: 0.799), our method substantially outperforms VAE, VQVAE, VQGAN, and f-AnoGAN baselines. Critically, cross-dataset evaluation on Srinivasan achieves AUROC 0.884 with superior generalization, demonstrating robust domain adaptation. On the external RETOUCH benchmark, unsupervised anomaly segmentation achieves competitive Dice (0.200) and mIoU (0.117) scores, validating reproducibility across institutions. 

\end{abstract}    
\section{Introduction}
\label{sec:intro}

OCT is a ubiquitous, noninvasive modality for retinal assessment in routine clinical practice, capturing micrometer-resolution layer morphology within seconds. Its widespread adoption, even in resource-limited settings, positions OCT as an ideal substrate for early detection and longitudinal monitoring of sight-threatening disease. While recent deep learning models have achieved strong performance on OCT classification and layer segmentation \cite{awais2017classification,wang2019oct,lu2019deep}, these systems are typically developed on large, internally curated datasets with predefined disease categories. Upon deployment to external cohorts, differences in imaging device, acquisition protocols, and patient populations frequently lead to significant performance degradation \cite{ZHANG2021e665}.

More fundamentally, supervised approaches are constrained by annotation availability: they can only recognize abnormalities explicitly labeled during training. Rare conditions, subtle structural deviations, or pathologies absent from the training set may be missed entirely. Additionally, acquiring high-quality clinical annotations is labor-intensive, costly, and difficult to scale—a critical barrier to deployment in diverse clinical environments. These limitations underscore the need for annotation-efficient learning paradigms that generalize robustly across heterogeneous clinical settings.

A complementary direction reframes retinal analysis as \emph{anomaly detection}: learn the manifold of healthy anatomy and flag deviations for detection and localization \cite{marimont2020vqvaeAD, pinaya2022generativeAD, seebock2019exploiting}. This approach is clinically compelling because (1) unlabeled normal OCTs are far more abundant than expertly labeled disease scans; (2) diverse case-mixes make exhaustive disease enumeration impractical; and (3) pixel-level anomaly signals enable direct localization and clinical triage. However, existing methods often operate in continuous feature spaces or provide coarse localization.

We investigate discrete latent modeling for OCT anomaly detection and localization. We train a Vector-Quantized Generative Adversarial Network (VQGAN) on normal macular B-scans to learn a compact codebook of healthy retinal patterns. Discretization naturally suppresses device artifacts and background variation, while providing interpretable control over reconstruction fidelity. Crucially, we introduce retinal layer-aware supervision and structured triplet learning to align learned representations with retinal anatomy and improve separation between healthy and pathological morphology. This yields a deployable model that reliably reconstructs healthy anatomy, detects anomalies via reconstruction discrepancies, and demonstrates strong cross-dataset generalization.

Our work demonstrates that anatomy-guided discrete latent modeling provides a practical, annotation-free foundation for OCT anomaly detection with strong cross-dataset generalization and clinical robustness. Quantitative evaluation demonstrates AUROC scores of 0.799 on the internal dataset and 0.8838 on the external dataset for anomaly detection, while enabling unsupervised anomaly localization on RETOUCH with a Dice score of 0.1998, outperforming all baselines across the evaluated metrics. Overall the main contributions of this work are: 

\begin{itemize}
    \item Unsupervised learning of healthy OCT distributions with retinal layer-aware guidance, enabling detection across internal and external datasets without disease labels.
    
    \item Reconstruction-based pixel-wise anomaly maps for label-free localization and quantitative evaluation on clinical benchmarks.
    
    \item Retinal layer-aware training strategy that improves robustness to device variation and enhances sensitivity to pathological deviations.
\end{itemize}
\section{Related Works}
\label{sec:related}
Unsupervised anomaly detection in medical imaging is commonly framed as a normative modeling problem, where models are trained exclusively on healthy data and abnormalities are identified as deviations from the learned distribution of normal anatomy. Early generative approaches formalized this paradigm using adversarial learning. 
\cite{DBLP:journals/corr/SchleglSWSL17, schlegl2019f} demonstrated that a GAN trained on healthy medical images can capture the manifold of normal structure, and anomalies can be quantified via reconstruction residuals and latent discrepancies. To improve inference efficiency, they further introduced an encoder to directly map inputs into the latent space, enabling fast anomaly scoring through combined image and feature space residuals. Extending adversarial reconstruction to volumetric contexts, 
\cite{li2019mad} incorporated multiple adjacent MRI slices to account for cross-slice anatomical continuity, highlighting the importance of contextual modeling in detecting subtle disease patterns.

Autoencoder based approaches constitute another major line of work in medical anomaly detection. These methods reconstruct healthy appearance and treat reconstruction errors as anomaly indicators. A systematic evaluation in 
\cite{baur2021autoencoders} analyzed several architectural variants and demonstrated that reconstruction based segmentation performance is sensitive to model design and data conditions. Reconstruction objectives were further refined in \cite{shvetsova2021anomaly}, which proposed improvements in training and abnormality scoring for complex high resolution medical images. Robust reconstruction strategies were explored in 
\cite{kascenas2022denoising}, showing that denoising objectives can enhance tumor detection performance compared to certain variational approaches. Variational formulations, including VAE based one class classifiers, extend this framework by modeling latent distributions of normal anatomy and identifying deviations via reconstruction likelihood or latent space discrepancies\cite{wijanarko2024tri}.

More recently, diffusion-based generative modeling has been introduced for normative anomaly detection. 
\cite{frotscher2023unsupervised} trains diffusion probabilistic models on healthy brain MRI and detects anomalies by analyzing deviations in denoising transitions, reflecting a broader shift toward richer generative processes for modeling anatomical variability.

Retinal OCT imaging presents domain specific challenges due to its layered anatomical structure and fine-grained pathological variations. Early OCT focused approaches such as 
\cite{seebock2018unsupervised} applied deep generative modeling to healthy scans and analyzed reconstruction residual maps for both anomaly detection and biomarker discovery. Moving beyond reconstruction error alone, 
\cite{seebock2019exploiting} leveraged epistemic uncertainty in segmentation models trained on healthy data as a proxy for structural deviation.

Structural priors have also been incorporated into OCT anomaly detection. 
\cite{zhou2020encoding} modeled structure and texture jointly to better capture retinal geometry. Translation based strategies such as 
\cite{wang2021weakly} employed CycleGAN-style domain mappings to infer lesion regions from translation residuals while reducing reliance on dense annotations. Feature level modeling was explored in 
\cite{das2022anomaly}, which combined deep feature extraction with sparse representation to quantify deviations from normal retinal patterns. Variational extensions tailored to OCT include 
\cite{zhou2023spatial}, which introduced spatial contextual constraints and attention correction mechanisms, and 
\cite{li2023self}, which integrated self-supervised learning with a variational autoencoder backbone to support anomaly detection, staging, and segmentation in retinal imaging.

Collectively, these works share a common principle: learning representations of healthy anatomical structure and identifying abnormalities as deviations from that learned distribution, whether via reconstruction residuals, latent likelihoods, uncertainty estimates, or feature reconstruction errors. While advances in generative modeling, structural priors, and representation learning have improved detection performance, anomaly detection in retinal OCT remains fundamentally dependent on how effectively normal retinal anatomy is modeled and how sensitively deviations from that structure are quantified. This motivates approaches that emphasize structurally informed, representation aware modeling of normal retinal patterns to enhance anomaly sensitivity while preserving anatomical fidelity.
\section{Methodology}
\label{sec:formatting}

The proposed methodology employs an unsupervised reconstruction based framework for detecting and localizing retinal abnormalities in OCT B-scans. A VQGAN is trained exclusively on healthy scans to model the distribution of normal retinal anatomy through a discrete latent representation that captures retina layers structure and texture. During inference, the model reconstructs healthy appearing retinas while inherently suppressing pathological features, and reconstruction discrepancies between the input and output are subsequently analyzed to identify anomalous scans and delineate regions of disease. 

\subsection{Dataset}

\textbf{Kermany OCT.} We use the Kermany OCT dataset \cite{kermany2018identifying} to study disease specific retinal morphology in macular B-scans. The dataset comprises over 84,495 images at approximately $500\times 750$ resolution, organized into four clinically relevant categories: Choroidal Neovascularization (CNV), Diabetic Macular Edema (DME), Drusen, and Normal. CNV and DME involve exudation and neovascular changes; Drusen denotes extracellular deposits associated with early age-related macular degeneration; the Normal class contains scans without visible pathology. Labels were assigned by ophthalmology experts, and the dataset has become a standard benchmark for OCT classification.

\textbf{Srinivasan OCT.} We use the Srinivasan OCT dataset \cite{srinivasan2014fully} to evaluate retina layers integrity and pathological alterations across major macular diseases. The dataset consists of spectral domain OCT B-scans acquired from subjects with Age-related Macular Degeneration (AMD), DME, and healthy retina. Each volume contains high resolution cross-sectional images centered on the macula, with typical image dimensions of approximately 
$512\times 1024$ pixels. All images were annotated and verified by ophthalmology specialists, and the dataset is widely used for benchmarking OCT based disease classification and segmentation.

\textbf{RETOUCH OCT.} We additionally utilize the RETOUCH dataset \cite{RETOUCH}, introduced as part of the Retinal OCT Fluid Detection and Segmentation Benchmark and Challenge. The dataset comprises 112 spectral domain OCT volumes acquired from three different devices (Cirrus, Spectralis, and Topcon), providing realistic inter device variability. Each volume is annotated at the pixel level for three clinically relevant fluid types: intraretinal fluid (IRF), subretinal fluid (SRF), and pigment epithelial detachment (PED). Due to the availability of dense segmentation masks, we employ RETOUCH specifically for anomaly segmentation and localization experiments rather than classification, enabling quantitative evaluation of Dice and IoU performance under multi-device domain shift.


\subsection{Model Training and Architecture}
To perform AD, we adopt an unsupervised learning paradigm that models the distribution of healthy retinal morphology from OCT B-scans. By learning the structural characteristic of normal retinas, the model can subsequently identify and localize deviations indicative of pathology.

In addition to distribution modeling, we train the model using a contrastive learning strategy composed of triplets: an anchor image, a positive image, and a negative image. The anchor image is a healthy OCT B-scan in our training dataset, while the positive image is another healthy scan acquired from a different patient, encouraging the model to learn patient-invariant representations of normal retinal anatomy. The negative image is constructed by perturbing the anchor image in a controlled manner to simulate pathological deviations. This design explicitly teaches the representation space to separate healthy morphology from disease-like deformations while remaining fully unsupervised with respect to real pathological labels.

The baseline VQGAN was trained exclusively on normal B-scans from the Kermany dataset, comprising 48,286 images for training, 1,007 for validation, and a 2019 image test set containing both normal and pathological cases (CNV, DME, and drusen). This design assumes that learning a discrete latent vocabulary of healthy retinal structures establishes a robust reference manifold against which deviations can be identified as potential anomalies. 

\begin{figure*}[t]
    \centering
    \includegraphics[width=\textwidth]{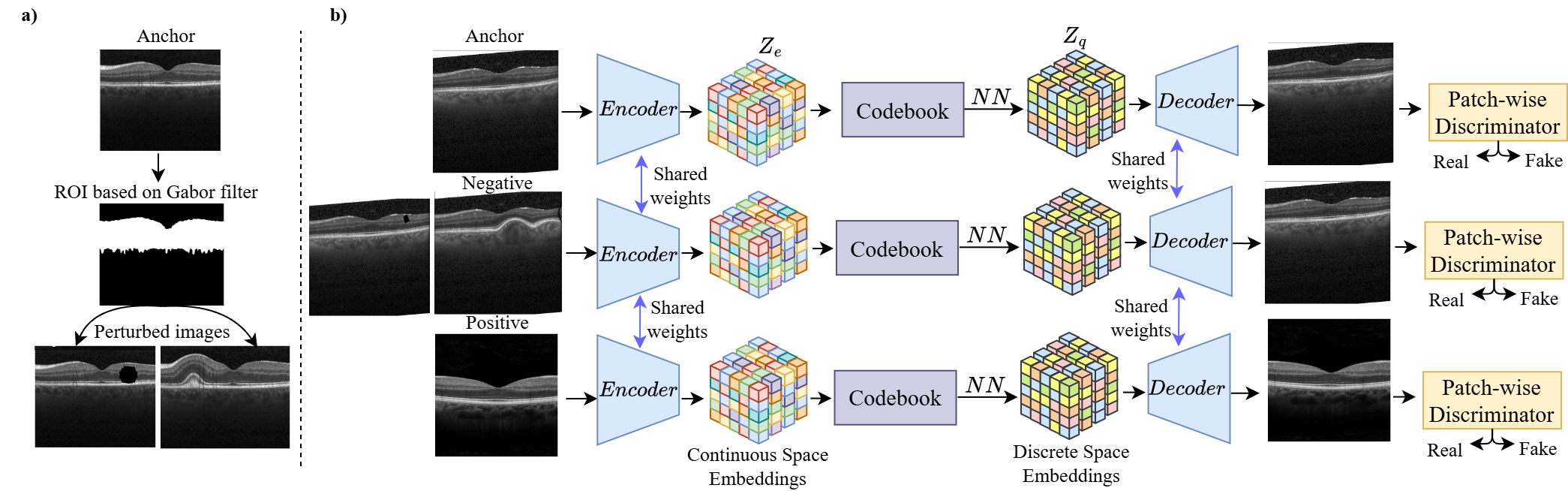}
    \caption{Overview of the proposed method. (a) Retina  layer extraction and generation of perturbed samples for triplet learning. (b) Discrete latent encoder–decoder framework with codebook quantization and patch-wise adversarial reconstruction.}
    \label{fig:framework}
\end{figure*}

Given an input retinal B-scan $x$, the VQGAN encoder $E(\cdot)$ first transforms the image into a compact latent representation $Z_e \in \mathbb{R}^{H' \times W' \times d}$, where each spatial position encodes a $d$-dimensional feature vector summarizing local retinal structure. Each latent vector is then quantized by replacing it with the nearest entry from a learnable codebook of $M$ discrete embeddings ${e_m}_{m=1}^M$. This produces a quantized latent map $Z_q$, constraining the representation space to a finite vocabulary of recurring anatomical patterns, such as layer boundaries and inter-layer textures. The decoder $D(\cdot)$ reconstructs an image $\hat{x} = D(Z_q)$, aiming to accurately reproduce the structural and textural details of the original retinal scan, as illustrated in ~\cref{fig:framework}.

Training optimizes a composite objective that balances pixel-level fidelity, perceptual consistency, adversarial realism, and latent-space discriminability. The reconstruction objective combines an $\ell_1$ pixel-wise loss with a learned perceptual similarity metric (LPIPS), thereby encouraging both low-level intensity accuracy and high-level structural coherence:

\begin{equation}
\mathcal{L}_{rec} =
\lambda_{1}\|x_t - \hat{x}\|_{1}
+
\lambda_{perc}\,\mathrm{LPIPS}(x_t,\hat{x})
\end{equation}

where $x_t$ denotes the designated reconstruction target.

To ensure stable and meaningful utilization of the discrete latent space, we employ the standard vector quantization objective consisting of embedding and commitment terms:

\begin{equation}
\mathcal{L}_{vq} =
\| \mathrm{sg}[z_e] - z_q \|_2^2
+
\beta \| z_e - \mathrm{sg}[z_q] \|_2^2
\end{equation}

where $z_e$ and $z_q$ represent the encoder outputs and their quantized counterparts, respectively, and $\mathrm{sg}[\cdot]$ denotes the stop-gradient operator.

To further prioritize anatomically meaningful structures, we introduce a region-of-interest (ROI) attention mechanism into the reconstruction objective. In retinal OCT B-scans, diagnostically relevant information is primarily concentrated within the retina layer. Therefore, instead of penalizing reconstruction errors uniformly across all pixels, we apply spatially adaptive weighting that increases the contribution of pixels inside the retinal region.
Formally, let $\mathrm{ROI}_i \in \{0,1\}$ denote a binary mask indicating whether pixel $i$ belongs to the retina layers. The ROI-weighted reconstruction loss is defined as:

\begin{equation}
\mathcal{L}_{ROI}
=
\frac{1}{N}
\sum_{i}
\left(1 + \alpha \cdot \mathrm{ROI}_i \right)
(x_i - \hat{x}_i)^2,
\end{equation}
where $\alpha$ controls the emphasis on retinal pixels. 
In our experiments, we set $\alpha = 6$, assigning a weight of $(1+ \alpha)$ to pixels within the retina layers while background pixels retain unit weight. 
This enforces anatomically guided reconstruction focused on clinically relevant retinal structures.

To promote structured separation in the latent representation space, we incorporate a triplet loss with a fixed margin of $m=1$:

\begin{equation}
\begin{aligned}
\mathcal{L}_{triplet}
=
\max \Big(
&\|f(x_a) - f(x_p)\|_2^2 \\
&- \|f(x_a) - f(x_n)\|_2^2
+ 1,\; 0
\Big)
\end{aligned}
\end{equation}

where $x_a$, $x_p$, and $x_n$ denote the anchor, positive, and negative samples, respectively, and $f(\cdot)$ represents the encoder embedding. This formulation encourages proximity between anchor–positive pairs while enforcing at least a unit margin of separation between anchor and negative representations.

Reconstruction targets are defined asymmetrically during training. When the input corresponds to either the anchor or the negative sample, the reconstruction target is set to the anchor image. In contrast, when the input corresponds to the positive sample, the reconstruction target is the positive image itself. This asymmetric formulation explicitly forces negative samples to be reconstructed toward the anchor (normal) image, effectively projecting anomalous inputs onto the learned normal data manifold. Meanwhile, positive samples are reconstructed to themselves, preserving identity consistency within the normal distribution.

Following a predefined warm-up phase (10,000 steps), a patch-based discriminator is introduced to encourage sharper and more realistic retinal reconstructions:

\begin{equation}
\mathcal{L}_{GAN}
=
\mathbb{E}[\log D(x_t)]
+
\mathbb{E}[\log (1 - D(\hat{x}))]
\end{equation}

The overall training objective is therefore given by:

\begin{equation}
\mathcal{L}_{total}
=
\mathcal{L}_{ROI}
+
\lambda_{perc}\mathcal{L}_{perc}
+
\lambda_{vq}\mathcal{L}_{vq}
+
\lambda_{triplet}\mathcal{L}_{triplet}
+
\lambda_{GAN}\mathcal{L}_{GAN}.
\end{equation}

This unsupervised optimization framework eliminates dependence on disease annotations while simultaneously encouraging perceptually faithful reconstruction, stable discrete representation learning, adversarial realism, structured latent separation, and projection of anomalous samples toward the learned normal manifold.

\begin{algorithm}[t]
\caption{Retina aware VQGAN training }
\label{alg:training}
\begin{algorithmic}[1]
\Require Normal OCT dataset $\mathcal{D}$ with patient IDs; VQGAN $(E,\{e_m\},G)$; warmup steps $T_w$
\For{each training iteration with mini-batch $\mathcal{B}$}
    \For{each anchor image $x_a \in \mathcal{B}$}
        \State Sample $x_p \sim \mathcal{D}$ s.t.\ $\text{pid}(x_p)\neq \text{pid}(x_a)$
        \State $\mathrm{ROI} \leftarrow \textsc{GaborROI}(x_a)$
        \State $x_n \leftarrow \textsc{Perturb}(x_a,\mathrm{ROI})$ \Comment{deform or noise}

        \For{$(x, x_t)\in \{(x_a,x_a),(x_p,x_p),(x_n,x_a)\}$}
            \State $z_e \leftarrow E(x)$;\;
            \State$z_q \leftarrow \textsc{Nearest}(z_e,\{e_m\})$;\;
            \State$\hat{x}\leftarrow G(z_q)$
            \State Compute $\mathcal{L}_{Recon}(x_t,\hat{x};\mathrm{ROI})+ \mathcal{L}_{vq}(z_e,z_q)$
        \EndFor

        \State Compute $\mathcal{L}_{triplet}\!\left(f(x_a),f(x_p),f(x_n)\right)$
        \If{step $> T_w$} \State Compute $\mathcal{L}_{GAN}$ \EndIf
    \EndFor
    \State Update model parameters using $\mathcal{L}_{total}$
\EndFor
\end{algorithmic}
\end{algorithm}

\textbf{Perturbation Strategy.} Negative samples are generated through a structured perturbation process confined to the retinal ROI, defined as the retina layers. To accurately localize this region, we first remove the white background artifacts introduced by dataset cropping procedures. A multi-scale horizontal Gabor filter \cite{li2023m} is applied to enhance the layered retinal texture. The maximum filter response is then binarized  to obtain an initial structural mask. Perturbations are subsequently applied within this anatomically constrained ROI to simulate pathological deviations. Specifically, we introduce two types of perturbations: (i) layer deformations that introduce curvature or local thickening of the retinal layers, and (ii)localized dark regions resembling intraretinal fluid accumulations.By restricting modifications to the retina layers, the generated negatives maintain global anatomical plausibility while embedding localized structural abnormalities, thereby encouraging the model to learn discriminative representations of healthy versus disease-like morphology, as summarized in Algorithm~\ref{alg:training}.

\textbf{Training Configuration.} Training is performed for 28 epochs with a learning rate of $1\times10^{-6}$, a codebook size of 256, codebook dimensionality of 256, batch size of 16, and input resolution of $256\times256$. After convergence, the decoder learns to reconstruct healthy appearing images conditioned on the quantized embeddings. At inference time, when the model processes a B-scan containing abnormalities, its reconstruction is naturally biased toward the normal manifold learned from healthy retinas. Consequently, pathological features such as intraretinal or subretinal fluid, pigment epithelial detachments, cystoid spaces, and other structural disruptions are often diminished or absent in the reconstructed image. The resulting differences between the input and its reconstruction highlight regions that deviate from normal retinal appearance, allowing for sensitive detection and precise localization of disease related abnormalities. 

\subsection{Evaluation Method}


To assess anomaly detection performance, we evaluate how effectively the model distinguishes pathological scans from the learned distribution of healthy retinal structure. We use L1 score for anomaly scoring and we use a weighted sum of L1 score and SSIM for spatial localization of anomalous regions. We intentionally decouple these two tasks: the former prioritizes stability and low false-positive rates, whereas the latter prioritizes spatial coherence and interpretability.



\paragraph{Anomaly localization.}

Reconstruction discrepancies can be leveraged to localize anomalous regions within an image. Various pixel-wise reconstruction metrics can be employed to generate error maps, including $\ell_1$, mean squared error (MSE), and structural similarity (SSIM). These discrepancy measures highlight regions where the reconstructed image deviates from the input, which often correspond to pathological structures. However, individual metrics may produce noisy or fragmented responses when used independently (see section~\ref{sec:appendix_metrics} for a detailed quantitative comparison).

In our experiments, we compute anomaly heatmaps using a weighted combination of intensity and structural discrepancies:
\begin{equation}
\label{eq:localization_metric}
\mathcal{E}(x, \hat{x}) 
=
\alpha\,|x - \hat{x}|
+
\beta\,(1 - \mathrm{SSIM}(x, \hat{x})),
\end{equation}
where $\alpha = 0.6$ and $\beta = 0.4$ in all experiments. This formulation balances pixel-level fidelity with structural consistency, yielding spatially coherent localization maps that better capture fluid regions and layer disruptions.

\begin{figure}[t]
    \centering
    \includegraphics[width=\linewidth]{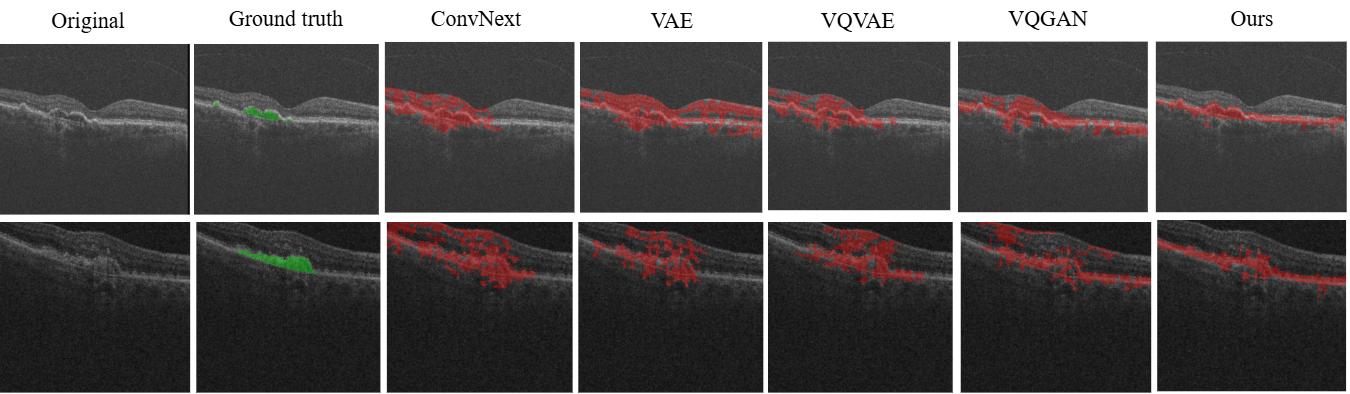}
    \caption{Qualitative comparison of anomaly segmentation results on OCT B-scans. Predicted anomaly regions are overlaid in red.}
    \label{fig:segmentation_comparison}
\end{figure}

Figure~\ref{fig:segmentation_comparison} presents qualitative segmentation maps obtained from different reconstruction based models using this error formulation, illustrating the spatial alignment between predicted anomaly regions and ground-truth annotations.

\section{Results}


We evaluate our framework across two complementary dimensions including anomaly detection performance, and anomaly localization capability. First, we analyze anomaly detection performance on both internal and external datasets to examine classification accuracy and cross-dataset generalization. Then, we evaluate unsupervised anomaly localization  to assess whether reconstruction discrepancies translate into spatially meaningful abnormality maps. Together, these experiments provide a comprehensive evaluation of generative fidelity, discriminative performance, and clinical relevance.

\subsection{Anomaly Detection}
\begin{figure}[t]
    \centering
    \includegraphics[width=\linewidth]{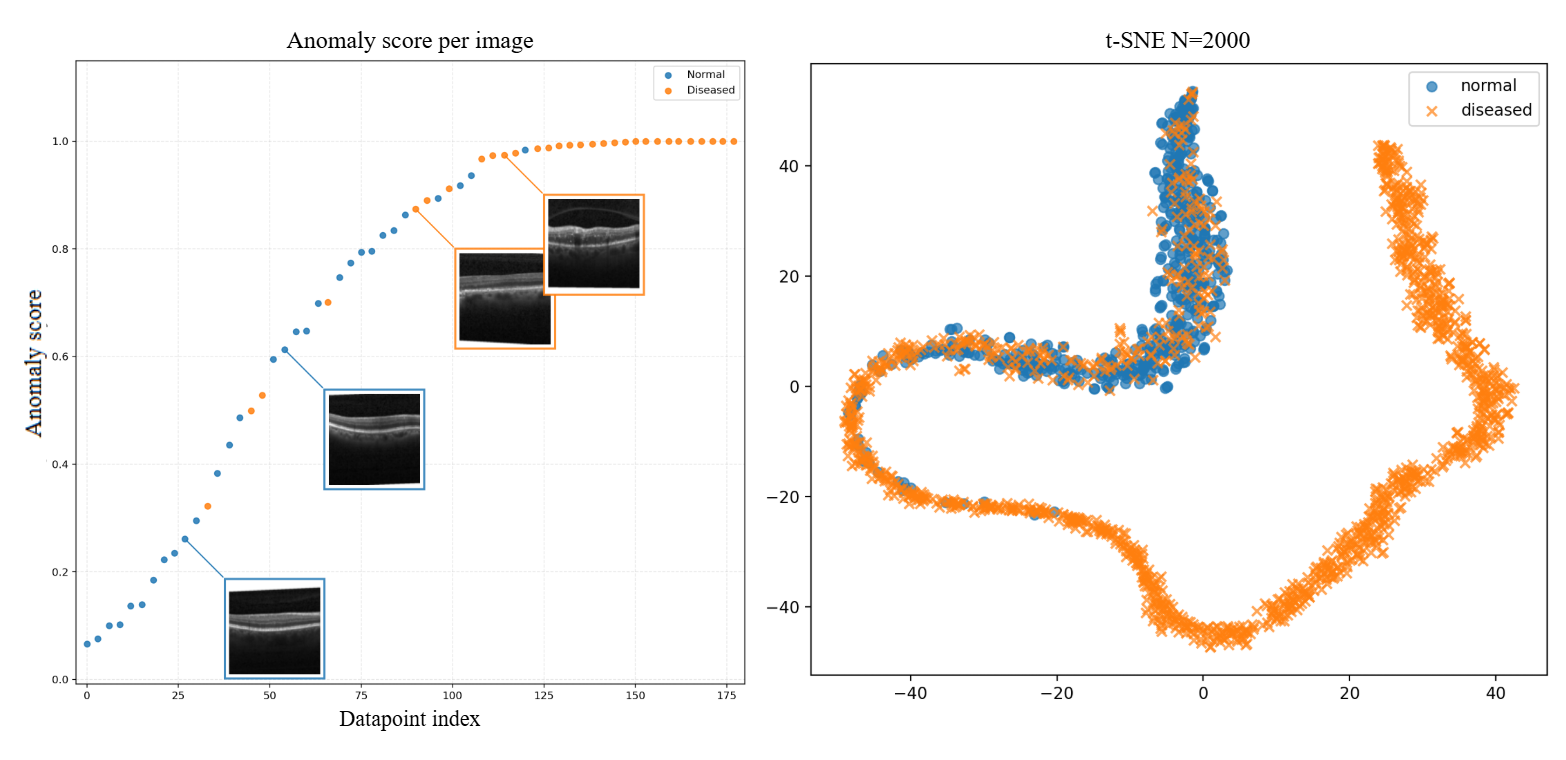}
    \caption{Left: Image-level anomaly scores sorted by datapoint index. Normal samples (blue) occupy lower score regions, while diseased samples (orange) exhibit progressively higher anomaly scores.  
    Right: t-SNE visualization ($N=2000$) of the learned latent embeddings, showing separation between normal and diseased samples in the learned representation space.}
    \label{fig:representation_analysis}
\end{figure}

To evaluate AD performance, we employed the L1 score to measure deviations from the learned distribution of healthy retinal structure. A fixed threshold $t^{*}$ was determined using the Youden $J$ statistic, which maximizes the balance between sensitivity and specificity on the validation set. Images with scores exceeding $t^{*}$ were classified as anomalous. The distribution of anomaly scores and the latent space separation between normal and diseased samples are illustrated in Fig.~\ref{fig:representation_analysis}.
 Classification metrics are summarized in ~\cref{tab:quantilegm_internal_external} for both the internal (Kermany) and external (Srinivasan) test sets. The external Srinivasan dataset comprises 400 images per condition (Normal, DME, and AMD), totaling 1,200 samples, providing an independent benchmark for cross-dataset generalization.

\textbf{Baseline Comparison.}
To rigorously evaluate the effectiveness of our framework, we compare against a diverse set of baselines spanning retinal foundation models, transformer and CNN based encoder–decoder architectures, continuous and discrete latent generative models, and OCT-specialized anomaly detection approaches. This design enables a controlled analysis of architectural inductive biases, latent space formulation, and adversarial learning.


We train a Vision Transformer (ViT-B)~\cite{dosovitskiy2020image} and a ConvNeXt~\cite{liu2022convnet} encoder, each followed by a linear projection decoder. We further compare against a Variational Autoencoder (VAE)~\cite{diederik2019introduction}, This baseline evaluates whether a smooth, continuous latent manifold is sufficient to model normal retinal structure and detect anomalies through reconstruction discrepancy. To investigate the impact of latent discretization, we include a Vector Quantized VAE (VQVAE)~\cite{van2017neural}, which enforces representational compactness through vector quantization and enables structured latent modeling without adversarial training. 

We also evaluate f-AnoGAN~\cite{schlegl2019f}, a GAN-based anomaly detection framework extensively adopted in medical imaging~\cite{wijanarko2024tri, tschuchnig2021anomaly, jang2021unsupervised, cheon2023sr} and specifically applied to OCT anomaly detection. f-AnoGAN models the distribution of normal anatomy and identifies abnormalities through reconstruction error and latent space consistency. As a clinically relevant and task aligned approach, it provides a strong specialized baseline for OCT anomaly detection.
 

\begin{table}[t]
\centering
\scriptsize
\setlength{\tabcolsep}{8pt}
\renewcommand{\arraystretch}{1.15}
\caption{Performance on internal and external test sets. Classification metrics are computed at thresholds $t^{*}$ determined by the Youden $J$ statistic. Best and second-best results for each metric are highlighted separately for internal and external validation.}

\begin{tabular}{lcccc}
\toprule
Model / configuration  & F1 & Acc & AUROC & $t^*$ \\
\midrule
\multicolumn{5}{c}{Internal validation} \\
\midrule

ViT-B 
& 66.36\% & 49.73\% & 0.4870 & 0.104739 \\

CONVNEXT 
& 69.66\% & 62.80\% & 0.7164 & 0.022586 \\

VAE 
& 69.60\% & 60.03\% & 0.6850 & 0.016798 \\

VQVAE
& \underline{71.36\%} & 65.53\% & 0.7365 & 0.021539 \\

VQGAN
& 71.31\% & 67.56\% & 0.7466 & 0.031344 \\

f-AnoGan
& 71.11\% & \underline{68.40\%} & \underline{0.7633} & 0.012371 \\

Ours
& \textbf{75.16\%} & \textbf{71.72\%} & \textbf{0.7990} & 0.032437 \\
\midrule

\multicolumn{5}{c}{External validation} \\
\midrule

ViT-B 
& \underline{80.0\%} & 66.67 & 0.2509 & - \\

CONVNEXT 
& 71.41\% & 69.37\% & 0.7767 & - \\

VAE 
& 77.60\% & 76.62\% & 0.8392 & - \\

VQVAE
& 78.13\% & \underline{78.37\%} & \underline{0.8457} & - \\

VQGAN
& 72.23\% & 67.13\% & 0.7726 & - \\

f-AnoGan
& 67.60\% & 53.50\% & 0.6794 & - \\

Ours
& \textbf{80.81\%} & \textbf{79.75\%} & \textbf{0.8838} & - \\
\midrule
\bottomrule
\end{tabular}
\label{tab:quantilegm_internal_external}
\end{table}

\subsection{Anomaly localization}
Our anomaly detection framework extends beyond image-level prediction to enable localization of anomalous regions in a fully unsupervised manner. Anomaly maps are generated from the pixel-wise difference between the reconstructed and original images, allowing abnormal regions to be highlighted without requiring pixel-level segmentation annotations during training. For quantitative evaluation, we use the validation split of the RETOUCH dataset, as the official test split does not provide publicly available ground-truth masks. Importantly, RETOUCH is treated as an external dataset for all compared models, and none of the methods were trained on it, ensuring a fair assessment of cross-dataset generalization. As shown in ~\cref{tab:unsupervised_segmentation}, we evaluate unsupervised anomaly segmentation performance using Dice and mean Intersection-over-Union (mIoU), which quantify the overlap between predicted anomaly regions and ground-truth masks.




\begin{table}[t]
\centering
\scriptsize
\setlength{\tabcolsep}{10pt}
\renewcommand{\arraystretch}{1.15}
\caption{Unsupervised anomaly segmentation performance measured using Dice and mean Intersection-over-Union (mIoU).}
\begin{tabular}{lcc}
\toprule
Model & Dice & mIoU \\
\midrule

VQGAN 
& 0.1449 & 0.0823 \\

ViT-B 
& 0.0039 & 0.0023 \\

ConvNext 
& 0.1813 & 0.1050 \\

VQVAE 
& \underline{0.1846} & \underline{0.1073} \\

VAE 
& 0.1234 & 0.0694 \\

Ours 
& \textbf{0.1998} & \textbf{0.1173} \\
\bottomrule
\end{tabular}
\label{tab:unsupervised_segmentation}
\end{table}






\subsection{Ablation Studies}
\textbf{Reconstruction Supervision and Positive Sample Selection.} To systematically assess the contribution of each design component, we conduct controlled ablation experiments by varying the reconstruction strategy and the definition of the positive sample in the triplet formulation. All experiments are trained under identical optimization and data conditions and evaluated using the same AD protocol to ensure a fair comparison. The quantitative results of these ablation experiments are summarized in ~\cref{tab:vqgan_ablations}.

In the first configuration, reconstruction loss is applied exclusively to the anchor image. The positive and negative samples are used only for computing the triplet loss and are not reconstructed by the decoder. This setting isolates the role of discriminative supervision in shaping the latent space, while the decoder focuses solely on modeling the healthy anchor distribution.

In the second configuration, reconstruction loss is applied to the anchor, positive, and negative samples. The positive image is selected as another healthy slice from the same patient as the anchor. This design leverages intra-subject anatomical consistency while preserving natural slice-level variations. By reconstructing all samples, the decoder is exposed to a richer set of structural patterns.

In the third configuration, reconstruction is again applied to all three samples, but the positive image is selected as a healthy slice from a different patient. This choice encourages the encoder to learn patient-invariant representations of healthy anatomy. Overall, the ablation study demonstrates that both the extent of reconstruction supervision and the definition of the positive sample play a critical role in AD performance.

\begin{table}[t]
\centering
\scriptsize
\setlength{\tabcolsep}{4pt}
\caption{VQGAN AD comparison. Best and second-best values per metric are highlighted.}
\begin{tabular}{lcccc}
\toprule
Model  & F1 & Acc & AUROC & t* \\
\midrule
tri-VQGAN (No Reconstruction Constraint)
 & \underline{0.730} & \underline{0.698} & \underline{0.760} & 0.031 \\

tri-VQGAN (Intra-Patient Positive Reconstruction)
 & 0.712 & 0.640 & 0.714 & 0.033 \\

tri-VQGAN (Inter-Patient Positive Reconstruction)
 & \textbf{0.738} & \textbf{0.699} & \textbf{0.774} & 0.032 \\
\bottomrule
\end{tabular}
\label{tab:vqgan_ablations}
\end{table}

\textbf{Objective Design.} 
We further analyze the contribution of each loss component through a progressive loss formulation study, summarized in Table~\ref{tab:loss_progression_ablation} for anomaly detection classification and in Table~\ref{tab:loss_progression_segmentation} for anomaly segmentation. Starting from the baseline reconstruction objective, we progressively add the triplet loss and subsequently the ROI attention loss. Each configuration is trained under identical settings, and the resulting performance is reported on both the internal and external anomaly detection datasets, as well as on RETOUCH for segmentation, to enable a direct comparison of different loss compositions across evaluation protocols. This controlled setup allows us to systematically examine the effect of incorporating additional supervision terms and to justify the final objective formulation.

\begin{table}[t]
\centering
\scriptsize
\setlength{\tabcolsep}{4pt}
\renewcommand{\arraystretch}{1.15}
\caption{Step-wise loss ablation on anomaly detection classification performance. Starting from the VQGAN reconstruction objective, triplet loss and ROI attention loss are progressively added. Results are reported on both internal and external datasets. Best and second-best values per metric are highlighted separately for each dataset.}
\begin{tabular}{lcccc}
\toprule
Loss Configuration & F1 & Acc & AUROC & $t^*$ \\
\midrule
\multicolumn{5}{c}{Internal Dataset} \\
\midrule
Reconstruction (VQGAN) 
& 0.7131 & 0.6756 & 0.746605 & 0.031344 \\

+ Triplet loss 
& \underline{0.7377} & \underline{0.6880} & \underline{0.774377} & 0.031623 \\

+ Triplet + ROI attention (Ours) 
& \textbf{0.7516} & \textbf{0.7172} & \textbf{0.799001} & 0.032437 \\

\midrule
\multicolumn{5}{c}{External Dataset} \\
\midrule
Reconstruction (VQGAN) 
& 0.7223 & 0.6713 & 0.772612 & 0.031592 \\

+ Triplet loss 
& \underline{0.7480} & \underline{0.7262} & \underline{0.823700} & 0.028713 \\

+ Triplet + ROI attention (Ours) 
& \textbf{0.8081} & \textbf{0.7975} & \textbf{0.883806} & 0.033595 \\
\bottomrule
\end{tabular}
\label{tab:loss_progression_ablation}
\end{table}

\begin{table}[t]
\centering
\scriptsize
\setlength{\tabcolsep}{6pt}
\renewcommand{\arraystretch}{1.15}
\caption{Unsupervised anomaly segmentation performance on the RETOUCH dataset (external). Starting from the VQGAN reconstruction objective, triplet loss and ROI attention loss are progressively added. Best and second-best values per metric are highlighted.}
\begin{tabular}{lcc}
\toprule
Loss Configuration & Dice & mIoU \\
\midrule
Reconstruction (VQGAN) 
& 0.1449 & 0.0823 \\

+ Triplet loss 
& \underline{0.1544} & \underline{0.0887} \\

+ Triplet + ROI attention (Ours) 
& \textbf{0.1998} & \textbf{0.1173} \\
\bottomrule
\end{tabular}
\label{tab:loss_progression_segmentation}
\end{table}

\section{Conclusion}

We proposed an unsupervised anomaly detection and localization framework for retinal OCT based on a discrete latent VQGAN trained exclusively on healthy B-scans. By combining vector quantization with retinal band-aware supervision and structured triplet learning, the model learns a compact normative representation of healthy anatomy and identifies pathology through reconstruction discrepancies, enabling both image-level detection and pixel-level localization without lesion annotations.

Across both internal and external evaluations, the proposed method achieves the highest AUROC among all compared models, demonstrating improved discriminative capability over transformer based encoders, continuous latent autoencoders, and GAN based anomaly detection approaches. On the internal dataset, our model surpasses strong generative baselines such as VAE, VQVAE, and VQGAN, and notably outperforms f-AnoGAN, which is a well-established generative anomaly detection framework that is frequently adopted as a reference baseline in medical imaging research, serving as a comparative standard for evaluating new anomaly detection models across diverse modalities and clinical applications. \cite{schlegl2019f, luo2023unsupervised, rahman2022healthygan, jebril2024anomaly, cai2025medianomaly, seebock2024anomaly}. While f-AnoGAN achieves strong AUROC internally (0.7633), our method improves this to 0.7990, indicating more effective separation between healthy and pathological samples in the learned representation space. The advantage becomes even more pronounced under cross-dataset evaluation: on the external dataset, our model achieves an AUROC of 0.8838, exceeding both reconstruction-based baselines and f-AnoGAN by a substantial margin. These results suggest that discrete latent modeling combined with anatomy-aware supervision yields a more stable and transferable representation of normal retinal structure, particularly under domain shift across devices and acquisition settings.

This cross-dataset generalization can be understood in light of how negative samples are constructed during training: rather than aiming to faithfully replicate real pathological manifestations, they are generated via heuristic perturbations and serve as a proxy supervision signal that encourages the model to learn sensitivity to deviations from normal anatomical structure. Consequently, the model is not constrained to recognize a predefined set of anomaly patterns, but rather to capture the underlying distribution of healthy retinal anatomy. This distinction is essential for generalization, as real-world pathological variations often differ substantially from synthetic corruptions. From this perspective, the discrepancy between generated perturbations and true abnormalities is not inherently detrimental; on the contrary, it may help mitigate overfitting to a limited set of artificial patterns. The strong performance observed on external datasets containing previously unseen pathologies further suggests that the learned representation encodes robust and generalizable anatomical priors, rather than merely memorizing characteristics of the synthetic negatives.

To further evaluate localization capability, we examine pixel-level segmentation performance, where the proposed method demonstrates consistent improvements over prior unsupervised approaches. However, the absolute metrics reported in \cref{tab:unsupervised_segmentation} remain relatively modest, which is expected given the fully unsupervised setting. The model is trained exclusively on healthy B-scans without pixel-level annotations and is not explicitly optimized for segmentation; instead, anomaly localization emerges from reconstruction discrepancies, enabling it to capture a broader range of structural irregularities beyond predefined annotated regions. Moreover, evaluation is performed on an external segmentation dataset not seen during training, introducing additional domain shift, yet the model still maintains performance gains over competing methods.

Despite these gains, several limitations should be acknowledged. The retinal band extraction and perturbation mechanisms rely on heuristic preprocessing and handcrafted transformations, which may not perfectly capture the diversity of real pathological patterns. Reconstruction-based anomaly scoring can also be sensitive to threshold selection and may under-detect abnormalities that are well reconstructed or overreact to acquisition artifacts. Furthermore, the current framework operates on 2D B-scans and is trained on a single healthy source distribution, leaving open questions about volumetric modeling, rare disease coverage, and broader multi-center validation.

In summary, this work demonstrates that anatomy-guided discrete latent modeling provides a robust and scalable foundation for unsupervised OCT anomaly detection. By explicitly modeling healthy retinal structure and enforcing structured latent separation, the proposed framework improves AUROC over established medical anomaly detection methods, including f-AnoGAN, and generalizes effectively across datasets, supporting its potential for real-world clinical screening and triage.

\section{Acknowledgement}
This work was supported by the U.S. National Institute of Health - National Eye Institute under Grants R15EY035804 and R21EY035271.

{
    \small
    \bibliographystyle{ieeenat_fullname}
    \bibliography{main}

@String(AAAI = {AAAI})

@article{van2017neural,
  title={Neural discrete representation learning},
  author={Van Den Oord, Aaron and Vinyals, Oriol and others},
  journal={Advances in neural information processing systems},
  volume={30},
  year={2017}
}

@article{dosovitskiy2020image,
  title={An image is worth 16x16 words: Transformers for image recognition at scale},
  author={Dosovitskiy, Alexey and Beyer, Lucas and Kolesnikov, Alexander and Weissenborn, Dirk and Zhai, Xiaohua and Unterthiner, Thomas and Dehghani, Mostafa and Minderer, Matthias and Heigold, Georg and Gelly, Sylvain and others},
  journal={arXiv preprint arXiv:2010.11929},
  year={2020}
}

@article{li2023m,
  title={M 2GF: Multi-Scale and Multi-Directional Gabor Filters for Image Edge Detection},
  author={Li, Yunhong and Bi, Yuandong and Zhang, Weichuan and Ren, Jie and Chen, Jinni},
  journal={Applied Sciences},
  volume={13},
  number={16},
  pages={9409},
  year={2023},
  publisher={MDPI}
}

@inproceedings{liu2022convnet,
  title={A convnet for the 2020s},
  author={Liu, Zhuang and Mao, Hanzi and Wu, Chao-Yuan and Feichtenhofer, Christoph and Darrell, Trevor and Xie, Saining},
  booktitle={Proceedings of the IEEE/CVF conference on computer vision and pattern recognition},
  pages={11976--11986},
  year={2022}
}

@inproceedings{cheon2023sr,
  title={SR-AnoGAN: you never detect alone. super resolution in anomaly detection (Student Abstract)},
  author={Cheon, Minjong},
  booktitle={Proceedings of the AAAI Conference on Artificial Intelligence},
  volume={37},
  number={13},
  pages={16194--16195},
  year={2023}
}

@article{jang2021unsupervised,
  title={Unsupervised anomaly detection using generative adversarial networks in 1H-MRS of the brain},
  author={Jang, Joon and Lee, Hyeong Hun and Park, Ji-Ae and Kim, Hyeonjin},
  journal={Journal of Magnetic Resonance},
  volume={325},
  pages={106936},
  year={2021},
  publisher={Elsevier}
}

@inproceedings{tschuchnig2021anomaly,
  title={Anomaly detection in medical imaging-a mini review},
  author={Tschuchnig, Maximilian E and Gadermayr, Michael},
  booktitle={International Data Science Conference},
  pages={33--38},
  year={2021},
  organization={Springer}
}

@inproceedings{wijanarko2024tri,
  title={Tri-vae: Triplet variational autoencoder for unsupervised anomaly detection in brain tumor mri},
  author={Wijanarko, Hansen and Calista, Evelyne and Chen, Li-Fen and Chen, Yong-Sheng},
  booktitle={Proceedings of the IEEE/CVF conference on computer vision and pattern recognition},
  pages={3930--3939},
  year={2024}
}

@article{schlegl2019f,
  title={f-AnoGAN: Fast unsupervised anomaly detection with generative adversarial networks},
  author={Schlegl, Thomas and Seeb{\"o}ck, Philipp and Waldstein, Sebastian M and Langs, Georg and Schmidt-Erfurth, Ursula},
  journal={Medical image analysis},
  volume={54},
  pages={30--44},
  year={2019},
  publisher={Elsevier}
}

@article{diederik2019introduction,
  title={An introduction to variational autoencoders},
  author={Diederik, P Kingma and Max, Welling},
  journal={Foundations and Trends{\textregistered} in Machine Learning},
  volume={12},
  number={4},
  pages={307--392},
  year={2019},
  publisher={Emerald Publishing Limited}
}

@article{RETOUCH,
	author={Hrvoje Bogunovi\'c and Freerk Venhuizen and Sophie Klimscha and Stefanos Apostolopoulos and Alireza Bab-Hadiashar and Ulas
	Bagci and Mirza Faisal Beg and Loza Bekalo and Qiang Chen and Carlos Ciller and Karthik Gopinath and Amirali K. Gostar and Kiwan
	Jeon and Zexuan Ji and Sung Ho Kang and Dara D. Koozekanani and Donghuan Lu and Dustin Morley and Keshab K. Parhi and
	Hyoung Suk Park and Abdolreza Rashno and Marinko Sarunic and Saad Shaikh and Jayanthi Sivaswamy and Ruwan Tennakoon and
	Shivin Yadav and Sandro De Zanet and Sebastian M. Waldstein and Bianca S. Gerendas and Caroline Klaver and Clara I. S\'anchez and
	Ursula Schmidt-Erfurth},
	title={{RETOUCH - The Retinal OCT Fluid Detection and
	Segmentation Benchmark and Challenge}},
	year={2019},
	journal = {IEEE Transactions on Medical Imaging},
	volume={38}, 
	number={8}, 
	pages={1858-1874}, 
	doi={10.1109/TMI.2019.2901398}, 
	ISSN={0278-0062}, 
	month={Aug}
}

@article{li2023self,
  title={Self-supervised anomaly detection, staging and segmentation for retinal images},
  author={Li, Yiyue and Lao, Qicheng and Kang, Qingbo and Jiang, Zekun and Du, Shiyi and Zhang, Shaoting and Li, Kang},
  journal={Medical Image Analysis},
  volume={87},
  pages={102805},
  year={2023},
  publisher={Elsevier}
}

@article{zhou2023spatial,
  title={Spatial--contextual variational autoencoder with attention correction for anomaly detection in retinal OCT images},
  author={Zhou, Xueying and Niu, Sijie and Li, Xiaohui and Zhao, Hui and Gao, Xizhan and Liu, Tingting and Dong, Jiwen},
  journal={Computers in biology and medicine},
  volume={152},
  pages={106328},
  year={2023},
  publisher={Elsevier}
}

@inproceedings{das2022anomaly,
  title={Anomaly detection in retinal images using multi-scale deep feature sparse coding},
  author={Das, Sourya Dipta and Dutta, Saikat and Shah, Nisarg A and Mahapatra, Dwarikanath and Ge, Zongyuan},
  booktitle={2022 IEEE 19th International Symposium on Biomedical Imaging (ISBI)},
  pages={1--5},
  year={2022},
  organization={IEEE}
}

@article{wang2021weakly,
  title={Weakly supervised anomaly segmentation in retinal OCT images using an adversarial learning approach},
  author={Wang, Jing and Li, Wanyue and Chen, Yiwei and Fang, Wangyi and Kong, Wen and He, Yi and Shi, Guohua},
  journal={Biomedical optics express},
  volume={12},
  number={8},
  pages={4713--4729},
  year={2021},
  publisher={Optical Society of America}
}

@inproceedings{zhou2020encoding,
  title={Encoding structure-texture relation with p-net for anomaly detection in retinal images},
  author={Zhou, Kang and Xiao, Yuting and Yang, Jianlong and Cheng, Jun and Liu, Wen and Luo, Weixin and Gu, Zaiwang and Liu, Jiang and Gao, Shenghua},
  booktitle={European conference on computer vision},
  pages={360--377},
  year={2020},
  organization={Springer}
}

@article{seebock2019exploiting,
  title={Exploiting epistemic uncertainty of anatomy segmentation for anomaly detection in retinal OCT},
  author={Seeb{\"o}ck, Philipp and Orlando, Jos{\'e} Ignacio and Schlegl, Thomas and Waldstein, Sebastian M and Bogunovi{\'c}, Hrvoje and Klimscha, Sophie and Langs, Georg and Schmidt-Erfurth, Ursula},
  journal={IEEE transactions on medical imaging},
  volume={39},
  number={1},
  pages={87--98},
  year={2019},
  publisher={IEEE}
}

@article{seebock2018unsupervised,
  title={Unsupervised identification of disease marker candidates in retinal OCT imaging data},
  author={Seeb{\"o}ck, Philipp and Waldstein, Sebastian M and Klimscha, Sophie and Bogunovic, Hrvoje and Schlegl, Thomas and Gerendas, Bianca S and Donner, Rene and Schmidt-Erfurth, Ursula and Langs, Georg},
  journal={IEEE transactions on medical imaging},
  volume={38},
  number={4},
  pages={1037--1047},
  year={2018},
  publisher={IEEE}
}

@article{frotscher2023unsupervised,
  title={Unsupervised anomaly detection using aggregated normative diffusion},
  author={Frotscher, Alexander and Kapoor, Jaivardhan and Wolfers, Thomas and Baumgartner, Christian F},
  journal={arXiv preprint arXiv:2312.01904},
  year={2023}
}

@inproceedings{kascenas2022denoising,
  title={Denoising autoencoders for unsupervised anomaly detection in brain MRI},
  author={Kascenas, Antanas and Pugeault, Nicolas and O’Neil, Alison Q},
  booktitle={International Conference on Medical Imaging with Deep Learning},
  pages={653--664},
  year={2022},
  organization={PMLR}
}

@article{shvetsova2021anomaly,
  title={Anomaly detection in medical imaging with deep perceptual autoencoders},
  author={Shvetsova, Nina and Bakker, Bart and Fedulova, Irina and Schulz, Heinrich and Dylov, Dmitry V},
  journal={IEEE Access},
  volume={9},
  pages={118571--118583},
  year={2021},
  publisher={IEEE}
}

@article{baur2021autoencoders,
  title={Autoencoders for unsupervised anomaly segmentation in brain MR images: a comparative study},
  author={Baur, Christoph and Denner, Stefan and Wiestler, Benedikt and Navab, Nassir and Albarqouni, Shadi},
  journal={Medical image analysis},
  volume={69},
  pages={101952},
  year={2021},
  publisher={Elsevier}
}

@article{DBLP:journals/corr/SchleglSWSL17,
  author       = {Thomas Schlegl and
                  Philipp Seeb{\"{o}}ck and
                  Sebastian M. Waldstein and
                  Ursula Schmidt{-}Erfurth and
                  Georg Langs},
  title        = {Unsupervised Anomaly Detection with Generative Adversarial Networks
                  to Guide Marker Discovery},
  journal      = {CoRR},
  volume       = {abs/1703.05921},
  year         = {2017},
  url          = {http://arxiv.org/abs/1703.05921},
  eprinttype    = {arXiv},
  eprint       = {1703.05921},
  bibsource    = {dblp computer science bibliography, https://dblp.org}
}

@inproceedings{li2019mad,
  title={MAD-GAN: Multivariate anomaly detection for time series data with generative adversarial networks},
  author={Li, Dan and Chen, Dacheng and Jin, Baihong and Shi, Lei and Goh, Jonathan and Ng, See-Kiong},
  booktitle={International conference on artificial neural networks},
  pages={703--716},
  year={2019},
  organization={Springer}
}

@article{seebock2024anomaly,
  title={Anomaly guided segmentation: Introducing semantic context for lesion segmentation in retinal OCT using weak context supervision from anomaly detection},
  author={Seeb{\"o}ck, Philipp and Orlando, Jos{\'e} Ignacio and Michl, Martin and Mai, Julia and Schmidt-Erfurth, Ursula and Bogunovi{\'c}, Hrvoje},
  journal={Medical Image Analysis},
  volume={93},
  pages={103104},
  year={2024},
  publisher={Elsevier}
}

@article{cai2025medianomaly,
  title={MedIAnomaly: A comparative study of anomaly detection in medical images},
  author={Cai, Yu and Zhang, Weiwen and Chen, Hao and Cheng, Kwang-Ting},
  journal={Medical Image Analysis},
  volume={102},
  pages={103500},
  year={2025},
  publisher={Elsevier}
}

@article{jebril2024anomaly,
  title={Anomaly detection in optical coherence tomography angiography (OCTA) with a vector-quantized variational auto-encoder (VQ-VAE)},
  author={Jebril, Hana and Eseng{\"o}n{\"u}l, Meltem and Bogunovi{\'c}, Hrvoje},
  journal={Bioengineering},
  volume={11},
  number={7},
  pages={682},
  year={2024},
  publisher={MDPI}
}

@inproceedings{rahman2022healthygan,
  title={Healthygan: Learning from unannotated medical images to detect anomalies associated with human disease},
  author={Rahman Siddiquee, Md Mahfuzur and Shah, Jay and Wu, Teresa and Chong, Catherine and Schwedt, Todd and Li, Baoxin},
  booktitle={International Workshop on Simulation and Synthesis in Medical Imaging},
  pages={43--54},
  year={2022},
  organization={Springer}
}

@article{luo2023unsupervised,
  title={Unsupervised anomaly detection in brain MRI: Learning abstract distribution from massive healthy brains},
  author={Luo, Guoting and Xie, Wei and Gao, Ronghui and Zheng, Tao and Chen, Lei and Sun, Huaiqiang},
  journal={Computers in biology and medicine},
  volume={154},
  pages={106610},
  year={2023},
  publisher={Elsevier}
}

@article{srinivasan2014fully,
  title={Fully automated detection of diabetic macular edema and dry age-related macular degeneration from optical coherence tomography images},
  author={Srinivasan, Pratul P and Kim, Leo A and Mettu, Priyatham S and Cousins, Scott W and Comer, Grant M and Izatt, Joseph A and Farsiu, Sina},
  journal={Biomedical optics express},
  volume={5},
  number={10},
  pages={3568--3577},
  year={2014},
  publisher={Optical Society of America}
}

@article{kermany2018identifying,
  title={Identifying medical diagnoses and treatable diseases by image-based deep learning},
  author={Kermany, Daniel S and Goldbaum, Michael and Cai, Wenjia and Valentim, Carolina CS and Liang, Huiying and Baxter, Sally L and McKeown, Alex and Yang, Ge and Wu, Xiaokang and Yan, Fangbing and others},
  journal={cell},
  volume={172},
  number={5},
  pages={1122--1131},
  year={2018},
  publisher={Elsevier}
}

@article{lu2019deep,
  title={Deep-learning based multiclass retinal fluid segmentation and detection in optical coherence tomography images using a fully convolutional neural network},
  author={Lu, Donghuan and Heisler, Morgan and Lee, Sieun and Ding, Gavin Weiguang and Navajas, Eduardo and Sarunic, Marinko V and Beg, Mirza Faisal},
  journal={Medical image analysis},
  volume={54},
  pages={100--110},
  year={2019},
  publisher={Elsevier}
}

@inproceedings{awais2017classification,
  title={Classification of sd-oct images using a deep learning approach},
  author={Awais, Muhammad and M{\"u}ller, Henning and Tang, Tong B and Meriaudeau, Fabrice},
  booktitle={2017 IEEE International Conference on Signal and Image Processing Applications (ICSIPA)},
  pages={489--492},
  year={2017},
  organization={IEEE}
}

@article{wang2019oct,
  title={On OCT image classification via deep learning},
  author={Wang, Depeng and Wang, Liejun},
  journal={IEEE Photonics Journal},
  volume={11},
  number={5},
  pages={1--14},
  year={2019},
  publisher={IEEE}
}

@article{ZHANG2021e665,
title = {Clinically relevant deep learning for detection and quantification of geographic atrophy from optical coherence tomography: a model development and external validation study},
journal = {The Lancet Digital Health},
volume = {3},
number = {10},
pages = {e665-e675},
year = {2021},
issn = {2589-7500},
doi = {https://doi.org/10.1016/S2589-7500(21)00134-5},
url = {https://www.sciencedirect.com/science/article/pii/S2589750021001345},
author = {Gongyu Zhang and Dun Jack Fu and Bart Liefers and Livia Faes and Sophie Glinton and Siegfried Wagner and Robbert Struyven and Nikolas Pontikos and Pearse A Keane and Konstantinos Balaskas}
}

@article{marimont2020vqvaeAD,
  author    = {Sergio Naval Marimont and Giacomo Tarroni},
  title     = {Anomaly detection through latent space restoration using vector-quantized variational autoencoders},
  journal   = {arXiv preprint arXiv:2012.06765},
  year      = {2020},
  url       = {https://arxiv.org/abs/2012.06765}
}

@article{pinaya2022generativeAD,
  author    = {W. H. L. Pinaya and others},
  title     = {Unsupervised brain imaging 3D anomaly detection and segmentation with transformers: The MOSAIC+ proposal},
  journal   = {Medical Image Analysis},
  year      = {2022},
  volume    = {79},
  pages     = {102410},
  doi       = {10.1016/j.media.2022.102410},
  url       = {https://www.sciencedirect.com/science/article/pii/S1361841522001220}
}
}

\newpage
\clearpage
\setcounter{page}{1}
\maketitlesupplementary

\section{Reconstruction Metric Analysis for Anomaly Segmentation}
\label{sec:appendix_metrics}
This appendix provides a detailed analysis of the impact of different reconstruction discrepancy measures on anomaly localization performance. While the main paper reports segmentation results using our weighted reconstruction formulation, here we compare several commonly used pixel-wise error metrics, including $\ell_1$, mean squared error (MSE), structural dissimilarity $(1 - \text{SSIM})$, and a weighted combination of $\ell_1$ and SSIM defined as
\[
\mathcal{E}(x,\hat{x}) = 0.6|x-\hat{x}| + 0.4(1-\mathrm{SSIM}(x,\hat{x})).
\]

Table~\ref{tab:seg_metrics_comparison} reports mean Dice and mIoU scores over all 327 RETOUCH test images with available ground-truth annotations. The comparison is conducted across multiple reconstruction-based architectures, including VQGAN, VQVAE, ConvNeXt, and different variants of our Tri-VQGAN model.

This analysis serves two purposes. First, it demonstrates how the choice of reconstruction discrepancy metric influences segmentation quality, highlighting the trade-off between pixel-level intensity differences and structural similarity cues. Second, it justifies the use of our weighted formulation by showing its consistent and competitive performance across model variants.

\begin{table}[t]
\centering
\scriptsize
\setlength{\tabcolsep}{4pt}
\renewcommand{\arraystretch}{1.1}
\caption{Segmentation performance (mean Dice / mIoU) over 327 RETOUCH images using different reconstruction error metrics. The weighted metric corresponds to $\mathcal{E}(x,\hat{x}) = 0.6|x-\hat{x}| + 0.4(1-\mathrm{SSIM})$.}
\label{tab:seg_metrics_comparison}
\begin{tabular}{lcccc}
\toprule
Model & Weighted & L1 & SSIM & MSE \\
\midrule

\midrule
\multicolumn{5}{c}{\textit{VQGAN}} \\
Dice  & \textbf{0.1449} & 0.1200 & 0.1003 & 0.1199 \\
mIoU  & \textbf{0.0823} & 0.0670 & 0.0575 & 0.0670 \\

\midrule
\multicolumn{5}{c}{\textit{VQVAE}} \\
Dice  & \textbf{0.1846} & 0.1574 & 0.1099 & 0.1573 \\
mIoU  & \textbf{0.1073} & 0.0925 & 0.0623 & 0.0924 \\

\midrule
\multicolumn{5}{c}{\textit{ConvNeXt}} \\
Dice  & \textbf{0.1846} & 0.1574 & 0.1099 & 0.1573 \\
mIoU  & \textbf{0.1073} & 0.0925 & 0.0623 & 0.0924 \\

\midrule
\multicolumn{5}{c}{\textit{Ours}} \\
Dice  & \textbf{0.1998} & 0.1795 & 0.1572 & 0.1786 \\
mIoU  & \textbf{0.1173} & 0.1058 & 0.0962 & 0.1052 \\

\bottomrule
\end{tabular}
\end{table}

\clearpage

\end{document}